\documentclass[11pt,letterpaper]{article}
\usepackage{emnlp2016}
\usepackage{times}
\usepackage{latexsym}

\makeatletter
\newcommand{\@BIBLABEL}{\@emptybiblabel}
\newcommand{\@emptybiblabel}[1]{}
\makeatother
\usepackage{scrextend}
\usepackage{hyperref}

\usepackage{booktabs}
\usepackage{multirow}
\usepackage{microtype}
\usepackage{soul}

\usepackage{mathrsfs, amsmath, amssymb}
\usepackage{graphicx}
\usepackage{tikz, pgfplots}
\usepackage{blindtext}
\usepackage{enumitem}
\usepackage{xspace,url,paralist}
\usepackage{etoolbox,siunitx}
\robustify\bfseries

\graphicspath{ {images/} }

\emnlpfinalcopy

\def\emnlppaperid{512}

\newcommand{\softmax}{\ensuremath{\operatorname{softmax}}\xspace}

\newcommand{\dropout}{\ensuremath{\operatorname{dropout}}\xspace}
\newcommand{\xentropy}{\ensuremath{\operatorname{CrossEntropy}}\xspace}
\newcommand{\M}{\ensuremath{\operatorname{M}}\xspace}
\newcommand{\distance}{\ensuremath{\operatorname{distance}}\xspace}
\newcommand{\filter}{\ensuremath{\operatorname{filter}}\xspace}
\newcommand{\Relu}{\ensuremath{\operatorname{ReLU}}\xspace}

\newcommand{\eqnref}[1]{Equation~\eqref{#1}\xspace}
\newcommand{\tabref}[1]{Table~\ref{#1}\xspace}
\newcommand{\figref}[1]{Figure~\ref{#1}\xspace}
\newcommand{\secref}[1]{Section~\ref{#1}\xspace}

\newcommand{\vect}[2][]{\ensuremath{\mathbf{#2}_{\mathrm{#1}}}\xspace}

\newcommand{\dataset}[1]{\textsf{#1}\xspace}
\newcommand{\MR}{\dataset{MR}}
\newcommand{\Subj}{\dataset{Subj}}
\newcommand{\CR}{\dataset{CR}}
\newcommand{\SST}{\dataset{SST}}

\title{Learning Robust Representations of Text}

\author{Yitong Li \qquad Trevor Cohn \qquad Timothy Baldwin \\
  Department of Computing and Information Systems\\
  The University of Melbourne, Australia \\
  {\tt yitongl4@student.unimelb.edu.au, \{tcohn,tbaldwin\}@unimelb.edu.au}
  }

\date{}

\begin{document}

\maketitle

\begin{abstract}
  Deep neural networks have achieved remarkable results across many
  language processing tasks, however these methods are highly sensitive to noise and adversarial attacks.
  We present a regularization based method for limiting network sensitivity to its
  inputs, inspired by ideas from computer vision, thus learning models
  that are more robust.
  Empirical evaluation over a range of sentiment datasets with a
  convolutional neural network shows that, compared to a baseline
  model and the dropout method, our method achieves superior
  performance over noisy inputs and out-of-domain
  data.\footnote{Implementation
    available at \url{https://github.com/lrank/Robust-Representation}.}
\end{abstract}

\section{Introduction}
\label{sec:intro}

Deep learning has achieved state-of-the-art results across a range of computer vision \cite{krizhevsky2012imagenet}, speech recognition \cite{graves2013speech} and natural language processing tasks \cite{bahdanau2014neural,kalchbrenner2014convolutional,yih2014semantic,bitvai2015non}.
However, 
deep models are often overconfident for noisy test instances, making them susceptible to adversarial attacks \cite{nguyen2015deep,tabacof2015exploring}.
\newcite{goodfellow2014explaining} argued that the primary cause of
neural networks' vulnerability to adversarial perturbation is their
linear nature, due to neural models being intentionally designed to
behave in a mostly linear manner to facilitate optimization.
\newcite{fawzi2015analysis} provided a theoretical framework for analyzing the robustness of classifiers to adversarial perturbations, and also showed linear models are usually not robust to adversarial noise.

In this work, we present a regularization method which makes deep learning models more robust to noise, inspired by \newcite{rifai2011contractive}.
The intuition behind the approach is to stabilize predictions by minimizing the ability of features to perturb predictions, based on high-order derivatives.
\newcite{rifai2011contractive} introduced contractive auto-encoders based on similar ideas, using the Frobenius norm of the Jacobian matrix as a penalty term to extract robust features.
Also related, \newcite{martens2010deep} investigated a second-order optimization method based on Hessian-free approach for training deep auto-encoders.
Where our proposed approach differs is that we train models using
first-order derivatives of the training loss as part of a
regularization term, necessitating second-order derivatives for
computing the gradient. 
We empirically demonstrate the effectiveness of the model over text
corpora with increasing amounts of artificial masking noise, using a range of sentiment analysis datasets \cite{pang2008opinion} with a convolutional neural network model \cite{kim2014convolutional}.
In this, we show that our method is superior to dropout
\cite{srivastava2014dropout} and a baseline method using MAP training.

\section{Training for Robustness}

Our method introduces a regularization term during training to ensure model robustness. 
%
%
We develop our approach based on a general class of parametric models, with the following structure. 
Let \vect{x} be the input, which is a sequence of (discrete) words,  represented by a fixed-size vector of continuous values, \vect{h}.
A transfer function takes \vect{h} as input and produces an output distribution, \vect[pred]{y}.
Training proceeds using stochastic gradient descent to minimize a loss function $L$, measuring the difference between \vect[pred]{y} and the truth \vect[true]{y}.


The purpose of our work is to learn neural models which are more robust to strange or invalid inputs.
When small perturbations are applied on \vect{x}, we want the prediction \vect[pred]{y} to remain stable.
Text can be highly variable, allowing for the same information to be conveyed with different word choice, different syntactic structures, typographical errors, stylistic changes, etc.
This is a particular problem in transfer learning scenarios such as domain adaptation, where the inputs in distinct domains are drawn from related, but different, distributions.
A good model should be robust to these kinds of small changes to the input, and produce reliable and stable predictions.

Next we discuss methods for learning models which are robust to variations in the input, before providing details of the neural network model used in our experimental evaluation.

\subsection{Conventional Regularization and Dropout}

Conventional methods for learning robust models include $l_1$ and $l_2$ regularization \cite{ng2004feature}, and dropout \cite{srivastava2014dropout}.
In fact, \newcite{wager2013dropout} showed that the dropout regularizer is first-order equivalent to an $l_2$ regularizer applied after scaling the features.
Dropout is also equivalent to ``Follow the Perturbed Leader'' (FPL) which perturbs exponential numbers of experts by noise and then predicts with the expert of minimum perturbed loss for online learning robustness \cite{van2014follow}.
Given its popularity in deep learning, we take dropout to be a strong baseline in our evaluation.

The key idea behind dropout is to randomly zero out units, along with
their connections, from the network during training, thus limiting
the extent of co-adaptation between units.
We apply dropout on the representation vector \vect{h}, denoted 
$\hat{\mathbf{h}} = \dropout_{\beta}(\vect{h})  $,
where $\beta$ is the dropout rate.
Similarly to our proposed method, training with dropout requires gradient based search for the minimizer of the loss $L$.

We also use dropout to generate noise in the test data as part of our experimental simulations, as we will discuss later.

\subsection{Robust Regularization}

Our method is inspired by the work on adversarial training in computer vision \cite{goodfellow2014explaining}.
In image recognition tasks, small distortions that are indiscernible to humans can significantly distort the predictions of neural networks \cite{szegedy2013intriguing}.
An intuitive explanation of our regularization method is, when noise is applied to the data, the variation of the output is kept lower than the noise.
We adapt this idea from \newcite{rifai2011contractive} and develop the Jacobian regularization method.

The proposed regularization method works as follows.
Conventional training seeks to minimise the difference between \vect[true]{y} and \vect[pred]{y}.
However, in order to make our model robust against noise, we also want to minimize the variation of the output when noise is applied to the input.
This is to say, when perturbations are applied to the input, there should be as little perturbation in the output as possible.
Formally, the perturbations of output can be written as
 $ \vect[y]{p} = \M(\vect{x} + \vect[x]{p}) - \M(\vect{x}) $,
where \vect{x} is the input, \vect[x]{p} is the vector of perturbations applied to \vect{x}, \M expresses the trained model, \vect[y]{p} is the vector of perturbations generated by the model, and the output distribution $\vect{y} = \M(\vect{x})$.
Therefore
 \begin{align*}
\lim_{\vect{p_x} \rightarrow \vect{0}} \vect{p_y}
& \!=\!  \lim_{\vect{p_x} \rightarrow \vect{0}} 
\! \big( \M(\vect{x} + \vect[x]{p}) -    \M(\vect{x}) \big) 
\! = \! \frac{\partial \vect{y}}{\partial \vect{x}}  \cdot \vect{p_x}, \\
\text{and} ~~
&  \distance\left(\lim_{\vect[x]{p} \rightarrow \vect{0}} \vect[y]{p}/\vect[x]{p}, \vect{0} \right) = \left\Vert \frac{\partial \vect{y}}{\partial \vect{x}} \right\Vert_F ~.
 \end{align*}
In other words, minimising local noise sensitivity is equivalent to minimising the Frobenius norm of
the Jacobean matrix of partial derivatives of the model outputs
\emph{wrt} its inputs.

To minimize the effect of perturbation noise, our method involves an additional term in the loss function, in the form of the derivative of loss $L$ with respect to hidden layer \vect{h}.
Note that while in principle we could consider robustness to perturbations in the input $\vect{x}$, the discrete nature of $\vect{x}$ adds additional mathematical complications, and thus we defer this setting for future work.
Combining the elements, the new loss function can be expressed as
\begin{equation}
 \mathscr{L} = L + \lambda \cdot \left\Vert \frac{\partial L}{\partial \vect{h}} \right\Vert _2 ,
\label{eq:robust}
\end{equation}
where $\lambda$ is a weight term, and \distance takes the form of the
$l_2$ norm.
The training objective in \eqnref{eq:robust} supports gradient
optimization, but note that it requires the calculation of
second-order derivatives of $L$ during back propagation, arising from the $\partial L / \partial \vect{h}$ term.
Henceforth we refer to this method as \textbf{robust regularization}.


\subsection{Convolutional Network}

For the purposes of this paper, we focus exclusively on convolutional
neural networks (CNNs), but stress that the method is compatible with
other neural architectures and other types of parametric models (not just deep neural networks).
The CNN used in this research is based on the model proposed by \newcite{kim2014convolutional}, and is outlined below.

Let $S$ be the sentence, consisting of $n$ words $\{ w_1, w_2, \cdots, w_n \}$.
A look-up table is applied to $S$, made up of word vectors $\vect[i]{e} \in \mathbb{R}^m$ corresponding to each word $w_i$, where $m$ is the word vector dimensionality.
Thus, sentence $S$ can be represented as a matrix $\vect[S]{E} \in \mathbb{R}^{m \times n}$ by concatenating the word vectors $\vect[S]{E} = \bigoplus_{i=1}^n  \vect[w_i]{e}$.

A convolutional layer combined with a number of wide convolutional filters is applied to \vect[S]{E}.
Specifically, the $k$-th convolutional filter operator $\filter_k$
involves a weight vector $\vect[k]{w} \in \mathbb{R}^{m \times t}$,
which works on every $t_k$-sized window of $\vect[S]{E}$, and is
accompanied by a bias term $b \in \mathbb{R}$.
The \filter operator is followed by the non-linear function
$\mathscr{F}$, a rectified linear unit, \Relu, followed by a max-pooling operation, to generate a hidden activation
$h_{k} = \operatorname{MaxPooling}(\mathscr{F}( \filter_k(\vect[S]{E}; \vect[k]{w}, b))$.
Multiple filters with different window sizes are used to learn different local properties of the sentence.
We concatenate all the hidden activations $h_k$ to form a hidden layer
\vect{h}, with size equal to the number of filters.
Details of parameter settings can be found in \secref{subsec:3.2}.

The feature vector \vect{h} is fed into a final \softmax layer with a
linear transform to generate a probability distribution over labels
$$\vect[pred]{y} = \softmax(\vect{w} \cdot \vect{h} + \vect{b}) \, ,$$ where \vect{w} and \vect{b} are parameters.
Finally, the model minimizes the loss of the cross-entropy between the ground-truth and the model prediction, $L = \xentropy(\vect[true]{y},\vect[pred]{y})$, for which we use stochastic gradient descent.

\begin{table*}[t]
  \footnotesize
  \centering
  \begin{tabular}{lll *{4}{S[round-mode=places,round-precision=1]} c *{4}{S[round-mode=places,round-precision=1]}}
  \toprule
  Dataset & &   & \multicolumn{4}{c}{\MR} &                       & \multicolumn{4}{c}{\Subj} \\
  \cmidrule{4-7} \cmidrule{9-12}
  Word dropout rate ($\alpha$) &&&   ${0}$ & 0.1 & 0.2 & 0.3 &                    & ${0}$ & 0.1 & 0.2 & 0.3 \\
  \midrule
  Baseline
  &           & & 80.45 & 79.41 & 77.87 & 76.52 &                 & \textbf{93.1} & 92.02 & 90.90 & 89.76 \\
  \midrule
  
  & $0.3$     & & 80.29 & 79.5 & 78.05 & 76.72 &                 & 92.70 & 92.01 & 90.89 & 89.53 \\
  Dropout ($\beta$)
  & $0.5$     & & 80.30 & 78.99 & 78.02 & 76.50 &                 & 93.0 & 92.03 & 91.1 & 89.87 \\ 
  & $0.7$     & & 80.29 & 79.26 & 78.33 & 76.83 &                 & 92.79 & 91.85 & 90.89 & 89.76 \\
  \midrule
  & $10^{-3}$ & & 80.47 & 79.41 & 78.34 & 76.69 &          & 93.0 & \textbf{92.2} & 91.1 & 89.77 \\
  Robust
  & $10^{-2}$ & & \textbf{80.8} & 79.25 & 78.4 & 77.0 &         & 93.0 & 92.17 & 91.01 & 90.0 \\
  Regularization ($\lambda$)
  & $10^{-1}$ & & 80.42 & 78.83 & 77.82 & 77.0 &        & 92.71 & 91.86 & 90.96 & 89.83 \\
  & $1$       & & 79.31 & 77.12 & 76.05 & 75.49 &                 & 91.66 & 91.07 & 90.14 & 89.31 \\
  \midrule
  Dropout + Robust & $\beta=0.5, \lambda=10^{-2}$
        & & 80.6 & \textbf{79.9} & \textbf{78.6} & \textbf{77.3} &  & 93.0 & \textbf{92.2} & \textbf{91.2} & \textbf{90.1} \\
  \toprule
  \\
  Dataset & &   & \multicolumn{4}{c}{\CR} &                       & \multicolumn{4}{c}{\SST} \\
  \cmidrule{4-7} \cmidrule{9-12}
  Word dropout rate ($\alpha$) &&&   ${0}$ & 0.1 & 0.2 & 0.3 &                    & ${0}$ & 0.1 & 0.2 & 0.3 \\
  \midrule
  Baseline
  &           & & 83.21 & 82.28 & 80.42 & 77.91 &                 & 84.12 & 82.32 & 80.28 & 77.81 \\
  \midrule
  
  & $0.3$     & & 83.25 & 82.06 & 80.26 & 78.90 &                 & 84.17 & 82.25 & 80.21 & 77.98 \\
  Dropout ($\beta$)
  & $0.5$     & & 83.23 & 82.43 & 80.95 & 79.28 &                 & 84.18 & 82.37 & 80.45 & 78.19 \\
  & $0.7$     & & 83.22 & 82.21 & 80.65 & 78.77 &                 & 83.90 & 82.48 & 80.90 & 78.19 \\
  \midrule
  & $10^{-3}$ & & 83.28 & 82.64 & 81.38 & 79.47 &                 & \textbf{84.5} & \textbf{82.8} & \textbf{81.4} & 78.8 \\
  Robust 
  & $10^{-2}$ & & \textbf{83.4} & 82.49 & 81.59 & 79.28 &        & 84.18 & 82.37 & 80.67 & 78.58 \\
  Regularization ($\lambda$)
  & $10^{-1}$ & & 83.29 & \textbf{82.7} & \textbf{82.0} & 79.6 &         & 82.53 & 81.49 & 79.73 & 77.59 \\
  & $1$       & & 82.89 & 81.35 & 79.76 & 79.04 &                 & 82.2 & 80.9 & 79.1 & 77.3 \\
  \midrule
  Dropout + Robust & $\beta=0.5, \lambda=10^{-2}$
              & & 83.3  & 82.5  & 81.5  & \textbf{79.7}  &        & 84.3 & 82.6 & 80.8 & \textbf{79.1} \\
  \bottomrule

  \end{tabular}
  \caption[T1]{Accuracy (\%) with increasing word-level dropout across the four datasets. For each dataset, we apply four levels of noise $\alpha = \{0, 0.1, 0.2, 0.3\}$; the best result for each combination of $\alpha$ and dataset is indicated in \textbf{bold}. The Baseline model is a simple CNN model without regularization. The last model combines dropout and our method with fixed parameters $\beta$ and $\lambda$ as indicated.}
  \label{tab:bt}
\end{table*}

\section{Datasets and Experimental Setups}

We experiment on the following datasets,\footnote{For datasets where there is no pre-defined training/test split, we evaluate using 10-fold cross validation. Refer to \newcite{kim2014convolutional} for more details on the datasets.} following \newcite{kim2014convolutional}:
\begin{compactitem}
    \item \MR: Sentence polarity dataset
      \cite{pang2008opinion}\footnote{\label{note1} \url{https://www.cs.cornell.edu/people/pabo/movie-review-data/}}
    \item \Subj: Subjectivity dataset \cite{pang2005seeing}\footref{note1}
    \item \CR: Customer review dataset \cite{hu2004mining}\footnote{\url{http://www.cs.uic.edu/~liub/FBS/sentiment-analysis.html}}
    \item \SST: Stanford Sentiment Treebank, using the 3-class configuration \cite{socher2013recursive}\footnote{\url{http://nlp.stanford.edu/sentiment/}}
\end{compactitem}
In each case, we evaluate using classification accuracy.

\subsection{Noisifying the Data}

Different to conventional evaluation, we corrupt the test data with
noise in order to evaluate the robustness of our model.
We assume that when dealing with short text such as Twitter posts, it is common to see unknown words due to typos, abbreviations and sociolinguistic marking of different types \cite{Han:Baldwin:2011a,Eisenstein:2013}.
To simulate this, we apply word-level dropout noise to each document,
by randomly replacing words by a unique sentinel symbol.\footnote{This
  was to avoid creating new $n$-grams which would occur when symbols
  are deleted from the input. Masking tokens instead results in
  partially masked $n$-grams as input to the convolutional filters.}
This is applied to each word with probability $\alpha \in \{0, 0.1, 0.2, 0.3\}$.

We also experimented with adding different levels of Gaussian noise to
the sentence embeddings \vect[S]{E}, but found the results to be
largely consistent with those for word dropout noise, and therefore we
have omitted these results from the paper.



To directly test the robustness under a more realistic setting, we
additionally perform cross-domain evaluation, where we train a model on one dataset and apply it to another.
For this, we use the pairing of \MR and \CR, where the first dataset
is based on movie reviews and the second on product reviews, but both
use the same label set.
Note that there is a significant domain shift between these corpora,
due to the very nature of the items reviewed.

\subsection{Word Vectors and Hyper-parameters}
\label{subsec:3.2}

To set the hyper-parameters of the CNN, we follow the guidelines of \newcite{zhang2015sensitivity}, setting word embeddings to $m=300$ dimensions and initialising based on word2vec pre-training \cite{mikolov2013distributed}.
Words not in the pre-trained vector table were initialized randomly by the uniform distribution $U([-0.25, 0.25)^{m})$.
The window sizes of filters ($t$) are set to $3, 4, 5$, with $128$ filters for each size, resulting in a hidden layer dimensionality of $384 = 128 \times 3$.
We use the Adam optimizer \cite{kingma2014adam} for training.


\section{Results and Discussions}

The results for word-level dropout noise are presented in \tabref{tab:bt}.
In general, increasing the word-level dropout noise leads to a drop in
accuracy for all four datasets, however the relative dropoff in
accuracy for Robust Regularization is less than for Word Dropout, and
in 15 out of 16 cases (four noise levels across the four datasets), our method achieves the best result.
Note that this includes the case of $\alpha = 0$, where the test data
is left in its original form,  which shows  that Robust Regularization is also an effective means of preventing overfitting in the model.

For each dataset, we also evaluated based on the combination of Word Dropout and Robust Regularization using the fixed parameters $\beta=0.5$ and $\lambda=10^{-2}$, which are overall the best individual settings.
The combined approach performs better than either individual method
for the highest noise levels tested across all datasets.
This indicates that Robust Regularization acts in a complementary way to Word
Dropout.

\begin{table}[t]
  \centering
  \footnotesize
  \begin{tabular}{l@{\,\,\,\,\,}l@{\,\,}c@{\,\,\,}c}
    \toprule
    Train/Test & & \MR/\CR & \CR/\MR\\
    \midrule
    Baseline                  &           & 67.5 & 61.0 \\
    \midrule
                              & 0.3       & 71.6 & 62.2  \\
    Dropout ($\beta$)           & 0.5       & 71.0 & 62.1  \\
                              & 0.7       & 70.9 & 62.0  \\
    \midrule
                              & $10^{-3}$ & 70.8 & 61.6  \\
    Robust                    & $10^{-2}$ & 71.1 & \textbf{62.5}  \\
    Regularization ($\lambda$)& $10^{-1}$ & \textbf{72.0} & 62.2  \\
                              & $1$       & 71.8 & 62.3  \\
    \midrule
    Dropout + Robust          & $\beta=0.5, \lambda=10^{-2}$ & \textbf{72.0} & 62.4 \\
    \bottomrule
  \end{tabular}
  \caption[T2]{Accuracy under cross-domain evaluation;  the best result for each dataset is indicated in \textbf{bold}. }
  \label{tab:cd}
\end{table}

\tabref{tab:cd} presents the results of the cross-domain experiment, whereby we train a model on \MR and test on \CR, and vice versa, to measure the robustness of the different regularization methods in a more real-world setting.
Once again, we see that our regularization method is superior to
word-level dropout and the baseline CNN, and the techniques combined
do very well,  consistent with our findings for synthetic noise.


\subsection{Running Time}
\begin{figure}[t]
  \scalebox{0.7}{\input{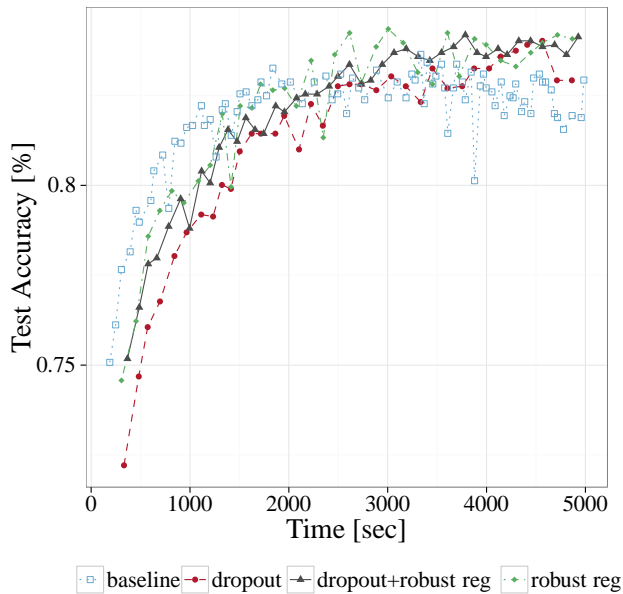}}
  \caption[T2]{Time--accuracy evaluation over the different
    combinations of Word Dropout (dropout) and Robust Regularization (robust reg) over \SST,
    without injecting noise.
  }
  \label{fig:time}
\end{figure}

Our method requires second-order derivatives, and thus is a little slower at training time.
\figref{fig:time} is a plot of the training and test accuracy at varying points during training over \SST.

We can see that the runtime till convergence is only slightly slower for Robust Regularization than standard training, at roughly 30 minutes on a two-core CPU (one fold) with standard training vs.\ 35--40 minutes with Robust Regularization.
The convergence time for Robust Regularization is comparable to that for Word Dropout.


\section{Conclusions}

In this paper, we present a robust regularization method which explicitly minimises a neural model's sensitivity to small changes in its hidden representation.
Based on evaluation over four sentiment analysis datasets using
convolutional neural networks, we found our method to be both superior
and complementary to conventional word-level dropout under varying levels of noise, and in a cross-domain evaluation.

For future work, we plan to apply our regularization method to other models and tasks to determine how generally applicable our method is.
Also, we will explore methods for more realistic linguistic noise,
such as lexical, syntactic and semantic noise, to develop models that
are robust to the kinds of data often encountered at test time.

\section*{Acknowledgments}

We are grateful to the anonymous reviewers for their helpful feedback and suggestions.
This work was supported by the Australian Research Council (grant
number FT130101105).
Also, we would like to thank the developers of Tensorflow
\cite{tensorflow2015-whitepaper}, which was used for the experiments in this paper.

\bibliography{emnlp2016}
\bibliographystyle{emnlp2016}

\end{document}